\documentclass[11pt,a4paper]{article}
\usepackage[hyperref]{emnlp2020}
\usepackage{times}
\usepackage{latexsym}

\usepackage{microtype}

\aclfinalcopy 

\usepackage{url}
\usepackage{graphicx}
\usepackage[font=small]{caption}
\usepackage{subcaption}
\usepackage{mathtools}
\usepackage{multirow}
\usepackage{amsmath}
\usepackage{enumitem}
\usepackage{tablefootnote}
\usepackage{threeparttable}
\setlist{noitemsep,topsep=0pt,parsep=0pt,partopsep=0pt,leftmargin=*}
\DeclareMathOperator*{\argmax}{arg\,max}

\usepackage{easyReview}

\usepackage{pgfplots}
\pgfplotsset{compat=1.15} 
\definecolor{forestgreen}{HTML}{009B55}
\definecolor{sepia}{HTML}{671800}
\definecolor{midnightblue}{HTML}{006795}
\definecolor{orangered}{HTML}{ED135A}

\title{Stepwise Extractive Summarization and Planning\\with Structured Transformers}

\author{Shashi Narayan\thanks{\quad Equal contribution.} \qquad \qquad \qquad Joshua Maynez$^\ast$ \qquad \qquad \qquad Jakub Adamek \\ 
\texttt{\{shashinarayan,joshuahm,enkait\}@google.com} 
\AND Daniele Pighin \qquad \qquad \qquad Bla\v{z} Bratani\v{c} \qquad \qquad \qquad Ryan McDonald \\ 
\texttt{\{biondo,blazb,ryanmcd\}@google.com} \AND
Google Research}

\date{}

\begin{document}
\maketitle

\begin{abstract}

We propose encoder-centric stepwise models for extractive summarization using structured transformers -- HiBERT \cite{zhang-etal-2019-hibert} and Extended Transformers \cite{etc}. We enable stepwise summarization by injecting the previously generated summary into the structured transformer as an auxiliary sub-structure. Our models are not only efficient in modeling the structure of long inputs, but they also
do not rely on task-specific redundancy-aware modeling, making them a 
general purpose extractive content planner for different tasks.
When evaluated on CNN/DailyMail extractive summarization, stepwise models achieve state-of-the-art performance in terms of Rouge without any redundancy aware modeling or sentence filtering. 
This also holds true for Rotowire table-to-text generation, where our models surpass previously reported metrics for content selection, planning and ordering, highlighting the strength of stepwise modeling.
Amongst the two structured transformers we test, stepwise Extended Transformers provides the best performance across both datasets and sets a new standard for these challenges.\footnote{The code and data are available at \url{https://github.com/google-research/google-research/tree/master/etcsum}.}

\end{abstract}

\section{Introduction}
\label{sec:intro}

Extractive document summarization is the task of creating a summary by identifying (and subsequently concatenating) the most important sentences in a document \cite{erkan2004lexrank,Nenkova:McKeown:2011}. In recent years this task has matured significantly, mostly thanks to advances in deep neural networks. 
\newcite{Cheng2016Neural} conceptualize extractive summarization as a sequence labeling task in which first a hierarchical long short-term memory network (LSTM; \citeauthor{Hochreiter1997Long}, \citeyear{Hochreiter1997Long}) is used to encode a document and then another LSTM is used to predict for each sentence whether it should be included in the summary. This architecture was later adopted by
\newcite{NallapatiZM16},
\newcite{Nallapati2017SummaRuNNer}, \newcite{narayan-etal-2018-ranking}, \newcite{zhang2018neural} and \newcite{dong-etal-2018-banditsum}.

Following the success of pre-trained transformer-based architectures for many tasks \cite{transformer,bert}, the current state-of-the-art approach to extractive summarization uses transformers to learn sentence representations and to rank sentences by their saliency \cite{bertextractive,Liu2019TextSW,zhang-etal-2019-hibert,zhong-etal-2019-searching,aredsum}. 
The top scoring sentences are then assembled to produce an extract of the document. Summaries built in this fashion \cite{Cheng2016Neural,narayan-sidenet18,zhang2018neural,dong-etal-2018-banditsum} are prone to contain redundant information. Several recent approaches have explored mechanisms to better handle redundancy, such as heuristic-based Trigram Blocking (TriBlk; \citeauthor{Liu2019TextSW}, \citeyear{Liu2019TextSW};  \citeauthor{wang-extsum-acl20}, \citeyear{wang-extsum-acl20}), handcrafted feature-driven models \cite{renetal2017} and redundancy aware neural sequence models \cite{zhou-etal-2018-neural,aredsum}. One common problem with these models is that their focus is limited to content overlap and to respecting length budgets. However, these are but a small subset of the dimensions necessary to produce informative and coherent summaries. Ideally, models would utilize enriched document and summary representations in order to implicitly learn better extractive plans for producing summaries \cite{nextsum,mendes-etal-2019-jointly}. One such method is \emph{stepwise} summarization \cite{nextsum}, where a summary is constructed incrementally by choosing new content conditioned on previously planned content.

In this paper, we propose encoder-centric stepwise models for extractive summarization using \emph{structured transformers}. Structured transformers are transformer-based architectures that have the flexibility to model some form of structure of the input, e.g., hierarchical document structure. In this paper, we specifically study two such architectures -- HiBERT \cite{zhang-etal-2019-hibert} and Extended Transformers Construction (ETC; \citeauthor{etc}, \citeyear{etc}). Details of these are given in Sections~\ref{sec:stepHiBERT} and~\ref{sec:stepetc}. We enable stepwise summarization by injecting the previously planned summary content into the structured transformer as an auxiliary sub-structure. 
The model then can holistically learn any document-level coherence properties, such as saliency, redundancy, and ordering, embodied in the gold summaries. This differs from other methods which are either task specific (e.g., redundancy aware modeling in \citeauthor{aredsum}, \citeyear{aredsum}) or not holistic (e.g., manually curated features in \citeauthor{nextsum}, \citeyear{nextsum}). An added advantage of structured encoders is that 
they break the quadratic attention mechanism of transformers \cite{bert}, making them more efficient and able to process longer inputs, instead of truncating the inputs to 512 tokens \cite{Liu2019TextSW,aredsum}, which is critical for long inputs and outputs which require non-trivial planning.  When evaluated on the CNN/DailyMail summarization dataset \cite{hermann-nips15}, we achieve state-of-the-art performance in terms of Rouge \cite{rouge} without any redundancy \cite{zhou-etal-2018-neural,aredsum} or sentence selection mechanisms \cite{Liu2019TextSW}.

Our model's task-agnostic approach allows it to implicitly learn and leverage content plans directly from the data. Moreover, structured transformers form the basis of our model, which are flexible in terms of content type (e.g., text or tables) that can be modeled. We demonstrate this by learning intricate extractive content plan for the Rotowire table-to-text generation task \cite{wiseman-etal-2017-challenges}. This task requires the generation of long summaries from large score tables detailing the
the specifics of a sports match, which
often necessitates dedicated content selection and planning models to generate a high-quality
summary \cite{wiseman-etal-2017-challenges,rotowire-ncpcc}. 
We show that our stepwise framework achieves higher content selection, planning and ordering scores relative to prior work with task-specific planning mechanisms.

The contributions of the paper are as follows: 
1) this is first study to use ETC \cite{etc} for summarization for its ability and flexibility to better model long and structured inputs; 2) we propose augmentions of two structured transformers, HiBERT and ETC, in order to enable stepwise models for extractive planning; 3) we demonstrate empirically that our models are general purpose and can be adapted as an extractive document summarizer or as a content planner for table-to-text generation; 4) Our experiments highlight the effectiveness of stepwise modeling, specifically stepwise ETC, which sets a new standard for both tasks.

\section{Related Work}
\label{sec:related}

\paragraph{Redundancy.}

Summarization models often use a dedicated {\em sentence selection} step after {\em sentence scoring} to address redundancy. Maximal Marginal Relevance \cite{mmr} based methods select the content that has the maximal score and is minimally redundant with the previously constructed partial summary. Others treated sentence selection as an optimization problem under some constraints such as summary length \cite{mcdonald07,lin-bilmes-2011-class}. \newcite{Liu2019TextSW} and \newcite{wang-extsum-acl20} used heuristic-based Trigram Blocking (TriBlk) for redundancy elimination. \newcite{renetal2017} trained two neural networks with handcrafted features;  one is used to rank sentences, and the other one is used to model redundancy during sentence selection. \newcite{zhou-etal-2018-neural} and \newcite{aredsum} proposed redundancy-aware models by modeling redundancy and saliency jointly during the scoring process using neural sequence models. In contrast to these approaches, our models are not redundancy-aware. Instead, they implicitly model redundancy by injecting previously generated summary representations. By virtue of this our models are not text-specific and can be applied to other tasks (see Section~\ref{sec:exprotowire}).

\paragraph{Partial Summary Representations.}

Ultilizing representations of partially generated summaries is relatively less studied in summarization. \newcite{mendes-etal-2019-jointly} proposed to dynamically model the generated summary using an LSTM to iteratively increment summaries based on previously extracted information. \newcite{nextsum} used a feed-forward neural network driven by hand-curated features capturing the prevalence of domain subtopics in the source and the summary. To the best of our knowledge, our models are first to use summary representations with structured transformers for summarization. Our models learn to make summary-informed next-sentence predictions without any hand-curated features.

\paragraph{Long-form Summarization.}

It is well known that a better content selection benefits abstractive summarizers to generate summaries that are not only fluent but also informative \cite{gehrmann2018bottom,Hsu2018Unified,copyrewrite-aaai20}. It can be particularly important when generating long abstractive summaries \cite{wikisum,liu-lapata-2019-hierarchical} or summarizing multiple documents \cite{Yasunaga2017Graph}.  Earlier multi-document summarization
methods have addressed the issue of long form input by graph-based representations of sentences or passages \cite{erkan2004lexrank,christensen-etal-2013-towards}. Recently, \newcite{Yasunaga2017Graph} proposed a neural version of this framework using graph convolutional networks \cite{Kipf2017SemiSupervisedCW}. \newcite{liu-lapata-2019-hierarchical} used cross-document attention mechanism to share information as opposed to simply concatenating text spans using hierarchical transformers. Similar to this motivation, we also explore better encoding of long inputs with structured transformers. 

\begin{figure*}[th!]
\centering
\begin{tabular}{ccc}
\includegraphics[scale=0.45]{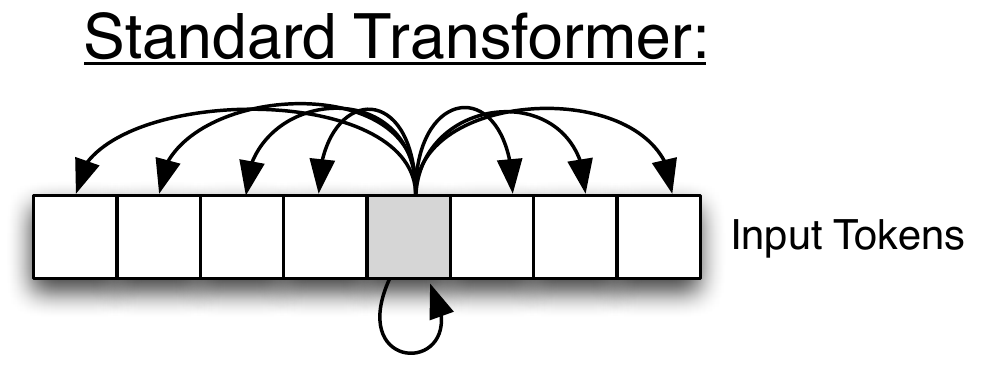} & 
\includegraphics[scale=0.45]{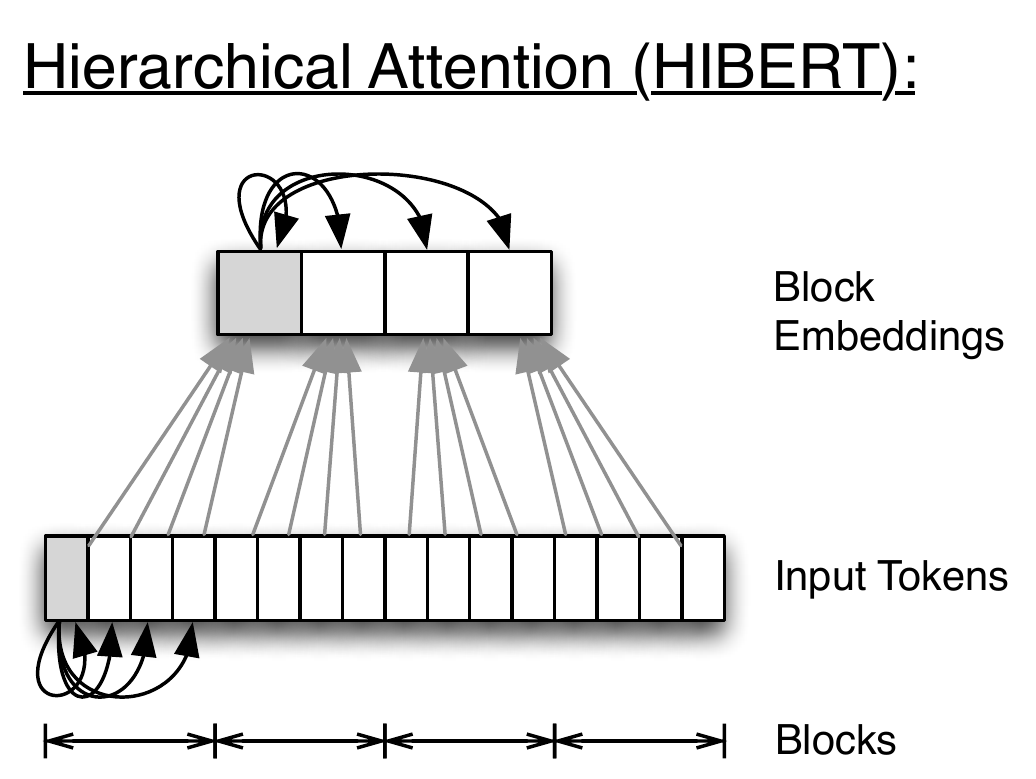} & 
\includegraphics[scale=0.10]{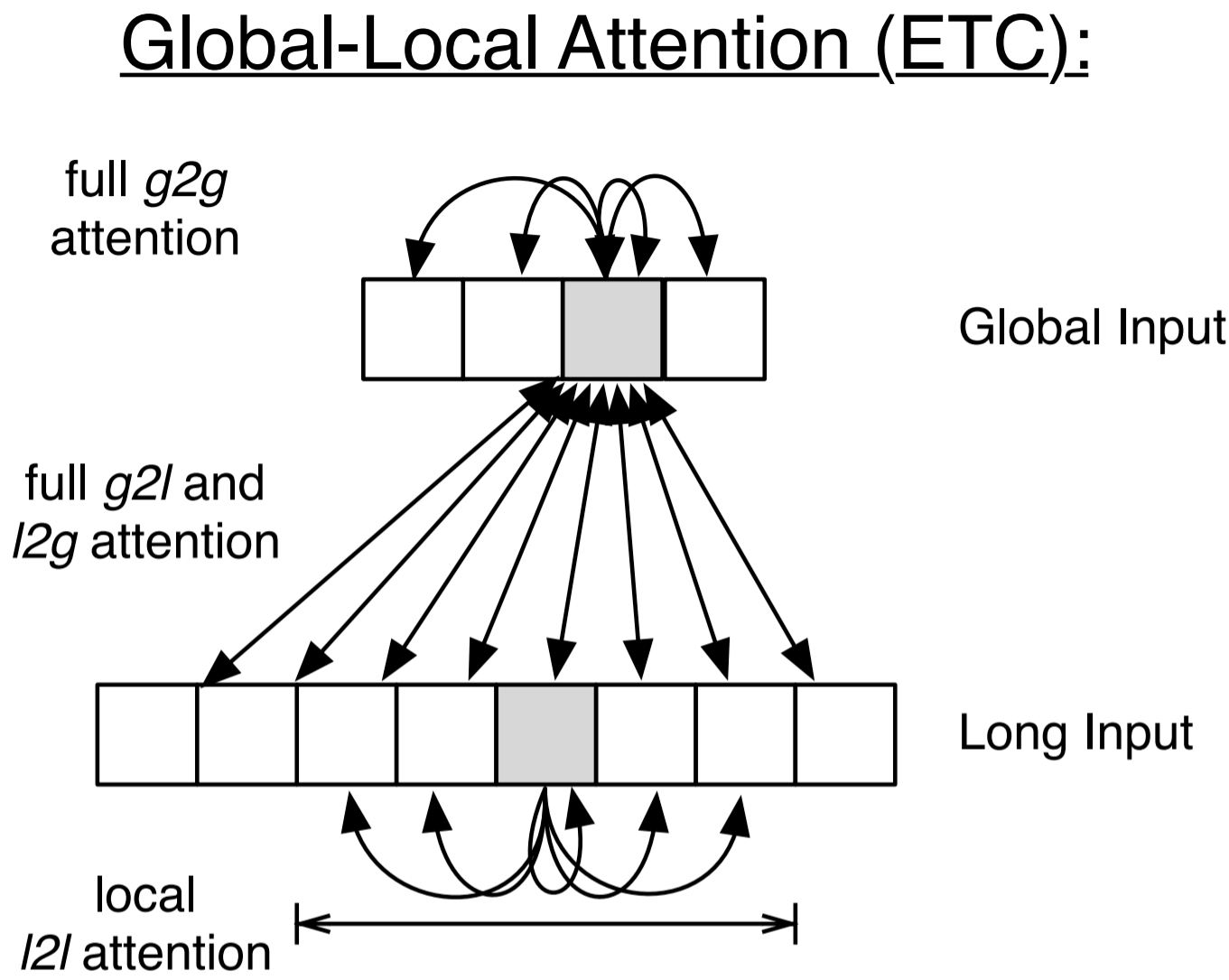}
\end{tabular}
\vspace{-0.2cm}
\caption{Memory usage and attentions in standard transformers \cite{bert}, HiBERT \cite{zhang-etal-2019-hibert} and ETC \cite{etc}.}
\label{fig:attention-rep}
\vspace{-0.4cm}
\end{figure*}

\paragraph{Table-to-Text Content Planning.}

\newcite{wiseman-etal-2017-challenges} introduced the Rotowire dataset, which requires multi-sentence summaries of large tables. Several works found that the key to generate fluent and informative summaries for this task is to have dedicated content planning and realization steps \cite{rotowire-ncpcc,rotowire-edinlg,rotowire-msgpt}. \newcite{rotowire-msgpt} and \newcite{rotowire-systran} used a transformer encoder, and, \newcite{gong-etal-2019-table} used multi-dimensional hierarchical LSTM encoders to compute better table entry representations. Following these lines of work, we evaluate our models to generate long content plans for this task using structured transformers.

\section{Problem: Stepwise Content Extraction}
\label{sec:problem} 

We define a general paradigm for stepwise content extraction that can be easily tailored to both extractive summarization and table-to-text generation. Given an input $D=\{s_1, s_2, \ldots, s_n\}$ with $n$ content units, the goal is to learn an extractive content plan, i.e., $S'_m=\{s'_j| 1 \leq j \leq m, s'_j \in (D \cup \{\O\})\}$, of length $m$; $s'_m$ is an empty unit ($\O$)  denoting the end of the plan. We formulate this as an iterative ranking problem \cite{nextsum,aredsum} where at each $k$-th step $(1 \leq k \leq m)$ given the input $D$ and the previously selected plan $S'_{k-1}$, we select $s'_k \in (D\cup \{\O\})$  with a probability $p(s'_k|S'_{k-1}, D; \theta)$ with model parameters $\theta$. The selected content is then added to $S'_{k-1}$ to construct $S'_k$. The best plan $\hat{S}$ can be defined as: 
\begin{center}
$\hat{S} = \argmax_{S'_m, \forall m} \prod_{k=1}^m P(s'_k|S'_{k-1}, D; \theta).$
\end{center}
For extractive document summarization, let $D=\{s_1, s_2, \ldots, s_n\}$ be a document with $n$ sentences. Our goal is to learn an extractive plan (or summary in this case) $\hat{S}$ which best summarizes $D$.
For table-to-text generation, we represent a table with $n$ records as $D=\{s_1, s_2, \ldots, s_n\}$. We aim to generate a  plan $S'_m$ that can be used by a text generator to generate a meaningful and coherent summary. 

For exposition, we use the extractive document summarization setup to introduce our stepwise models with HiBERT \cite{zhang-etal-2019-hibert} and ETC \cite{etc} in the following sections. Specifically, we use `sentence' as a content unit and `previously' or `partially generated summary' for a previously selected content plan.  

\section{Stepwise HiBERT}
\label{sec:stepHiBERT}

\begin{figure*}[th!]
\centering
\begin{tabular}{cc}
\includegraphics[scale=0.30]{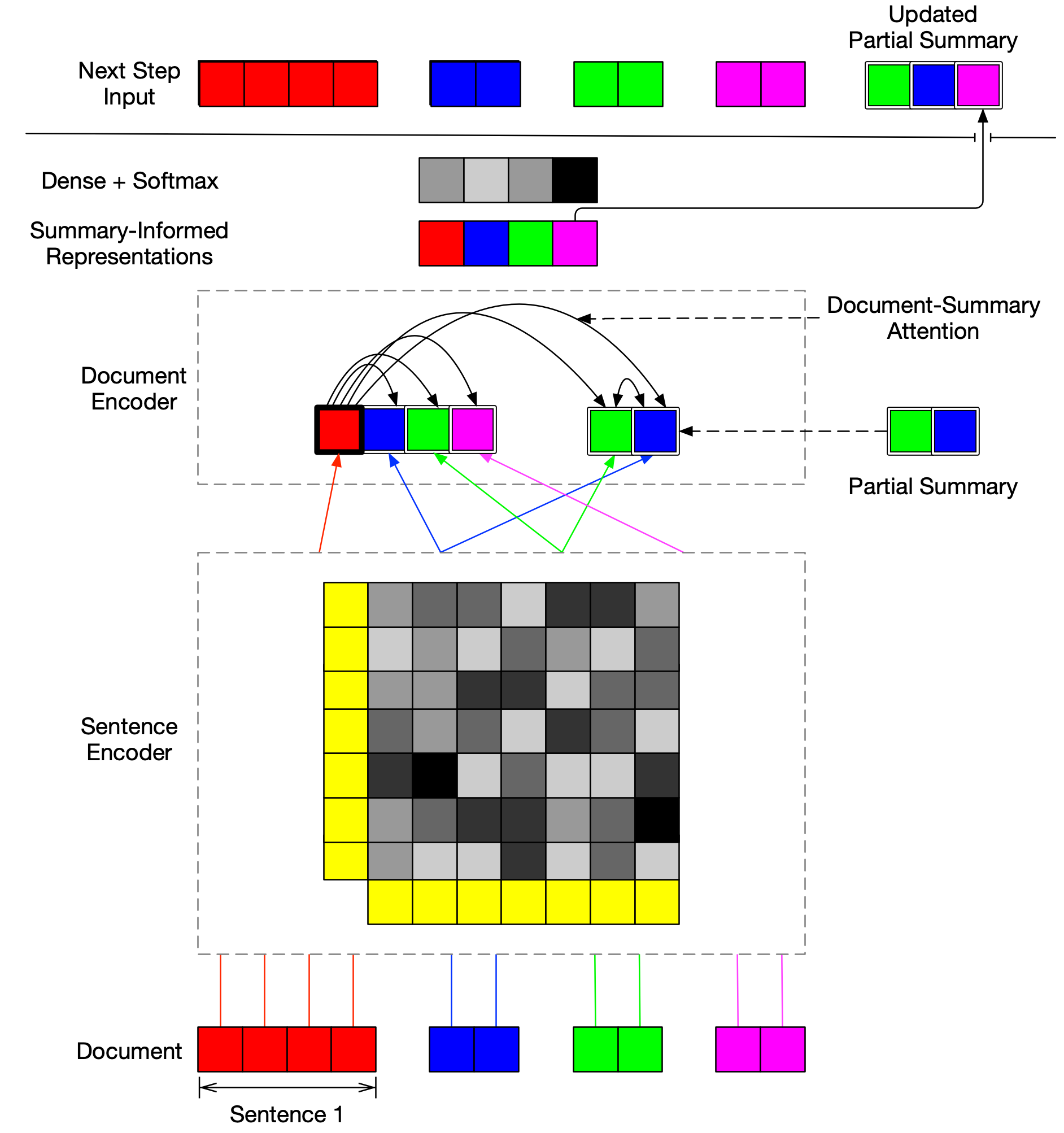} &
\includegraphics[scale=0.30]{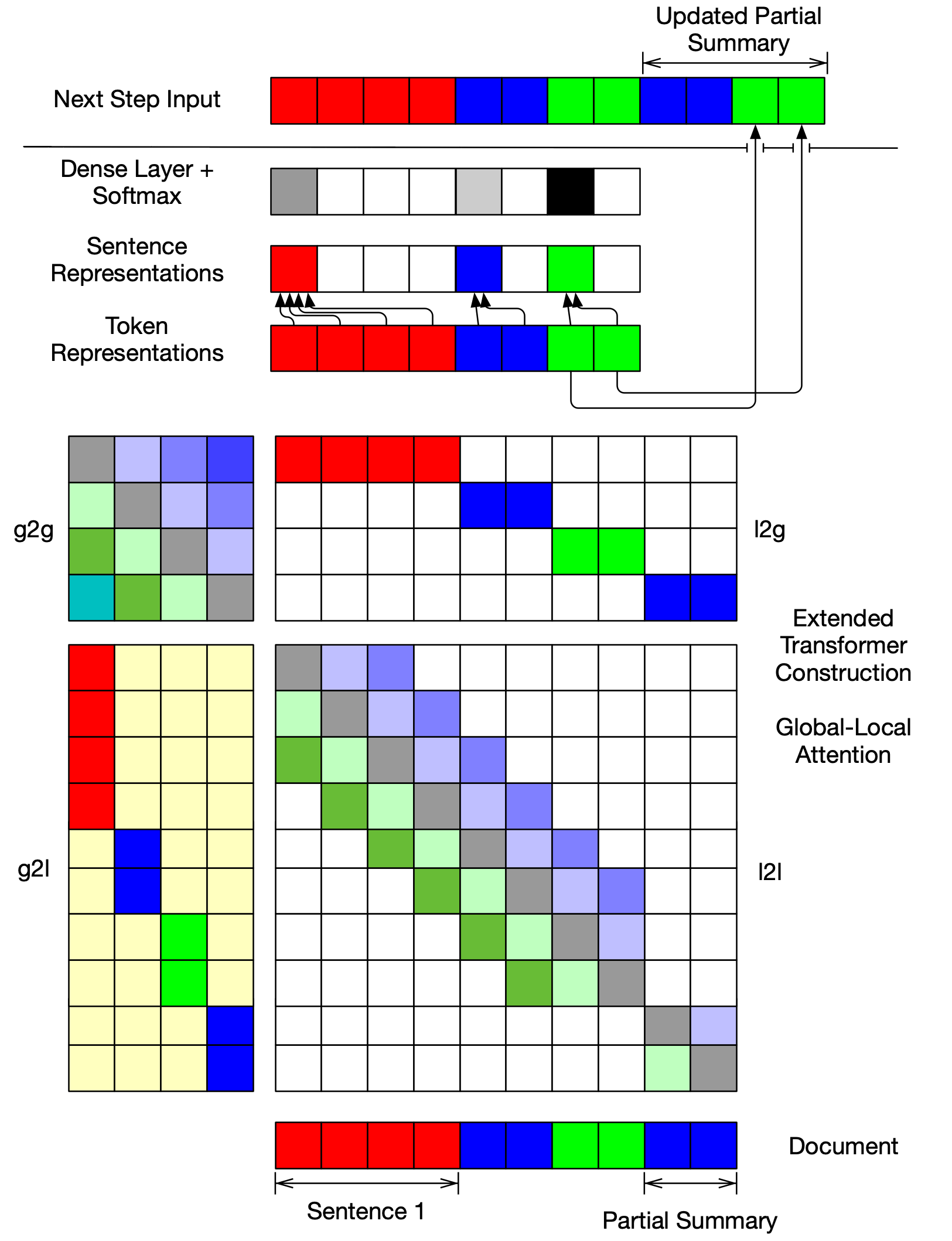} 
\end{tabular}
\vspace{-0.3cm}
\caption{Stepwise HiBERT (left) and ETCSum (right) models. HiBERT builds summary informed representation by jointly modeling partially generated summary and the document during document encoding, while ETCSum takes as input the document appended with the partially generated summary.}
\label{fig:stepwise}
\vspace{-0.45cm}
\end{figure*}

Hierarchical encodings have been used to  model input structure with LSTMs \cite{Nallapati2016Abstractive,Cheng2016Neural,narayan-etal-2018-ranking}.
\newcite{zhang-etal-2019-hibert} proposed \textbf{HiBERT} with two stacked Transformer encoders \cite{transformer} for extractive summarization (see the middle diagram in Figure~\ref{fig:attention-rep}): a {\em sentence encoder} that {\em independently} builds representations for each sentence in the document; and a {\em document encoder} that operates over sentence encodings to build contextual representations for all sentences. These contextual sentence representations are then ingested by a classifier to predict the salience score of each sentence in the document. As in standard transformers, both encoders have multiple layers with each layer composed of a multi-head self-attention layer followed by a feed-forward sub-layer with residual connections \cite{residual} and layer normalizations \cite{Ba2016LayerN}. For \textbf{Stepwise HiBERT}, at time step $k$, we modify the document encoder with the content plan $S'_{k-1}$, which is the previously selected sentences in the summary. This is depicted in Figure~\ref{fig:stepwise} (left) and allows the model to implicitly select new sentences relative to the previously generated summary.

\paragraph{Sentence and Document Encoders.}

Let $D=\{s_1, s_2, \ldots, s_n\}$ be a document, where $s_i=\{w^i_1, w^i_2, \ldots, w^i_{|s_i|}\}$ is a sentence in $D$ and $w^i_j$ is a token in $s_i$. $s_i$ is first mapped to a continuous space $\mathbf{E}_{s_i} = \{\mathbf{e}^i_1, \mathbf{e}^i_2, \ldots, \mathbf{e}^i_{|s_i|}\}$ where $\mathbf{e}^i_j = \mathbf{e}(w^i_j) + \mathbf{p}^{\text{token}}_j$. $\mathbf{e}(w^i_j)$ and $\mathbf{p}^{\text{token}}_j$ are the token and positional embeddings of token $w^i_j$, respectively. Our Transformer-based sentence encoder then transforms $\mathbf{E}_{s_i}$ into a list of hidden representations $\{\mathbf{h}^i_1, \mathbf{h}^i_2, \ldots, \mathbf{h}^i_{|s_i|}\}$, where $\mathbf{h}^i_j$ is the hidden representation for $w^i_j$. Following the standard practice \cite{bert,Liu2019TextSW}, we take the first hidden representation $\mathbf{h}^i_1$ as the representation for the sentence $s_i$.

\newcite{zhang-etal-2019-hibert} use a standard Transformer document encoder. It takes the document representation  $\hat{\mathbf{H}}_D=\{\mathbf{\hat{h}}^1, \mathbf{\hat{h}}^2, \ldots, \mathbf{\hat{h}}^n\}$, where $\mathbf{\hat{h}}^i = \mathbf{h}^i_1 + \mathbf{p}^{\text{sent}}_i$. $\mathbf{h}^i_1$ and $\mathbf{p}^{\text{sent}}_i$ are the representation from the sentence encoder and the positional embedding for sentence $s_i$ in the document, respectively, and, builds contextual sentence representations $\{\mathbf{d}_1, \mathbf{d}_2, \ldots, \mathbf{d}_n\}$. 

\paragraph{Stepwise Modeling.} 

At step $k$, let $S'_{k-1}=\{s'_1, s'_2, \ldots, s'_{k-1}\}$ be the partial summary with $(k-1)$ previously extracted sentences. In addition to $\hat{\mathbf{H}}_D$, our document encoder takes the summary representation $\hat{\mathbf{H}}_{S'_{k-1}}=\{\mathbf{\hat{x}}^1, \mathbf{\hat{x}}^2, \ldots, \mathbf{\hat{x}}^{k-1}\}$, where $\mathbf{\hat{x}}^i = \mathbf{h}^i_1 + \mathbf{p}^{\text{sum}}_i$. $\mathbf{h}^i_1$ is the representation from the sentence encoder for sentence $s_i$ and $\mathbf{p}^{\text{sum}}_i$ is the positional embedding for sentence $s_i$ in $S'_{k-1}$. At each layer, the document encoder employs three levels of nested multi-headed attentions \cite{transformer} to build summary-informed contextual sentence representations $\{\mathbf{d'}_1, \mathbf{d'}_2, \ldots, \mathbf{d'}_n\}$: {\em document self-attention}, {\em summary self-attention} and {\em document-summary attention} (see Figure~\ref{fig:stepwise}, left). The first two operate in parallel, followed by the document-summary attention.

While document self-attention learns the contextual hidden representation $\mathbf{h}^{\text{doc}\rightarrow\text{doc}}_{s_i}$ of each sentence in the document $D$, summary self-attention learns the contextual hidden representation $\mathbf{h}^{\text{sum}\rightarrow\text{sum}}_{s'_i}$ of each sentence in $S'_{k-1}$. We share the parameters of the document and summary self-attention layers. The document-summary attention then builds the contextual hidden representation $\mathbf{h}^{\text{doc}\rightarrow\text{sum}}_{s_i}$ of each sentence in the document $D$ using  linear projections of $\mathbf{h}^{\text{doc}\rightarrow\text{doc}}_{s_i}$ as query, and $\mathbf{h}^{\text{sum}\rightarrow\text{sum}}_{s'_i}$ as key and values \cite{transformer}.

In addition to the introduction of stepwise mechanism to HiBERT, our positional embeddings, $\mathbf{p}^{\text{token}}_j$, $\mathbf{p}^{\text{doc}}_j$ and $\mathbf{p}^{\text{sum}}_j$, are not shared to better model individual sentences, the document and the different styles of summary. \newcite{zhang-etal-2019-hibert} shared their token ($\mathbf{p}^{\text{token}}_j$) and sentence ($\mathbf{p}^{\text{sent}}_j$) positional embeddings. But we both use the absolute position encodings used in the original BERT model \cite{bert}. 


\section{Stepwise ETCSum}
\label{sec:stepetc}

There has been growing interest in addressing the limitation of the transformer architecture used in BERT \cite{bert} where memory usage scales quadratically with the size of the input \cite{guo-etal-2019-star,dai-etal-2019-transformer,Ye2019BPTransformerML,Child2019GeneratingLS,Rae2020CompressiveTF,Beltagy2020LongformerTL,Roy2020EfficientCS}. HiBERT alleviates this problem by modeling each sentence independently; the memory usage in HiBERT scales with the square of the number of sentences, and the square of the maximum length of any sentence. However, the main disadvantage of this approach is that token-level attention across sentences is prohibited and long range attention only happens indirectly at the second-stage encoder (see the middle diagram in Figure~\ref{fig:attention-rep}). Recently, Extended Transformer Construction (ETC; \citeauthor{etc}, \citeyear{etc}) provides an alternative. It alleviates the quadratic memory growth by introducing sparsity to the attention mechanism via its novel global-local attention mechanism (see the rightmost diagram in Figure~\ref{fig:attention-rep}). This not only permits encoding of long inputs,\footnote{As do other recent architectures \cite{xlnet_arxiv19,reformer}.} but also enables a mechanism to model structure directly through nodes in the global attention layer.

\paragraph{Global-Local Attention.}

The ETC model architecture receives two inputs: a long input, which in most cases corresponds to the text to be encoded; and an auxiliary global input, which serves as inductive bias features. First, the model builds an attention map, called {\em long-to-long}, across the long input with a sparse local attention of fixed length, this bypasses the quadratic memory complexity
and allows to scale input lengths to the thousands of tokens, but limits the attention span of tokens to their nearest neighbors.

To overcome this limitation, the global-local attention defines three other attention parts: {\em global-to-global}, {\em global-to-long} and {\em long-to-global}, all with unrestricted attention. This allows tokens arbitrarily far apart to attend to each other with at most one hop through the global input tokens. 
We refer the reader to \newcite{etc} for more details. The right parts of Figures~\ref{fig:attention-rep} and ~\ref{fig:stepwise} illustrate these four types of attentions and the sparsity diagrams where each cell in a row $i$ and column $j$ is different than white input token $w_i$ can attend to input token $w_j$, same relative position embeddings are indicated by using the same color.

\paragraph{Stepwise Modeling.}

Given the document $D$ and its partial summary $S'_{k-1}$ at step $k$, 
we construct an input $I=D^{\frown}S'_{k-1}=\{w_1, \ldots, w_{|D^{\frown}S'_{k-1}|}\}$ by concatenating the document $D$ and the partial summary $S'_{k-1}$. ETC replaces absolute position encodings with relative position encodings \cite{shaw-etal-2018-self} to easily adapt to greater input lengths than seen during pretraining. In addition to modeling relative positions in an input sequence, relative position encodings in ETC are also used to model arbitrary pairwise token relations useful for structured inputs.

We used the auxiliary global input to represent sentence structure. Specifically, following \cite{etc}, we placed one auxiliary token in the global input per each sentence in the input $I$. We linked the global tokens with the input tokens by using relative position labels to represent whether each token belongs to that sentence. Global-to-global attention is left unrestricted, allowing all sentences to attend to each other.
This result is summary-informed contextualized input token representations via attention through the global nodes. In the rest of the paper we refer to this summarizer by \textbf{Stepwise ETCSum}.
Similar to HiBERT, we take the first token hidden representation $\mathbf{h}^i_1$ as the representation for the sentence $s_i$. Finally, sentence embeddings are passed to the softmax layer for salience scoring. Both HiBERT and ETCSum are then trained with the cross entropy loss.

\section{Extractive Document Summarization}
\label{sec:expcnndm}
\subsection{Experimental Setup} 

\paragraph{Dataset.}
We evaluate our models on the CNN and DailyMail news highlights datasets \cite{hermann-nips15}. We used
standard splits (287,227/13,368/11,490 documents) for training, validation, and testing. We did not anonymize entities or lower case tokens as in \cite{narayan-etal-2018-ranking,zhou-etal-2018-neural,zhang-etal-2019-hibert,Liu2019TextSW}. The documents in the CNN/DailyMail dataset are long; the average lengths are 760.5 words (34 sentences) for CNN and 653.3 words (29.3 sentences), for DailyMail. The human written abstracts have 46 and 55 words for CNN and DailyMail, respectively. We evaluated summarization quality  using F$_1$ Rouge.\footnote{We lowercased candidate and reference summaries and used \texttt{pyrouge} with parameters ``-a -c 95 -m -n 4 -w 1.2.'' }

\paragraph{Baselines.} 

We compared our Stepwise HiBERT and ETCSum models to Lead and Oracle baselines. Lead selects the first 3 sentences to form the summary, while Oracle baselines creates a  summary by selecting the best possible set of sentences in the document that gives the highest average of Rouge-1, Rouge-2 and Rouge-L F1 scores with respect to the human written summary. The Oracle (512) truncates the input document to 512 tokens. We further compared our models against several redundancy-aware models (NeuSum; \citeauthor{zhou-etal-2018-neural}, \citeyear{zhou-etal-2018-neural}  and ARedSum; \citeauthor{aredsum}, \citeyear{aredsum}) and models that uses Trigram Blocking (TriBlk; \citeauthor{Liu2019TextSW}, \citeyear{Liu2019TextSW}) for redundancy elimination during sentence selection (see the second block in Table~\ref{tab:expcnndm}). 

To understand the importance of modeling long documents for extractive summarization, we also trained BERTSum, similar to \newcite{Liu2019TextSW}, with a receptive capacity of 512 tokens initialized with the BERT checkpoint. Our BERTSum differs slightly from \newcite{Liu2019TextSW}, in that we don't use segment embeddings. We also report on RoBERTa \cite{roberta} initialized version of BERTSum (RoBERTaSum). 

\begin{table}[t!]
\centering
\footnotesize
\begin{tabular}{ l|ccc}
\hline
Models & R1 & R2 & RL\\
\hline
Lead  & 40.42 & 17.62 & 36.67 \\
Oracle (512) & 52.59 & 31.24 & 48.87 \\
Oracle (Full) & 57.82 & 35.05 & 53.99  \\ \hline
Latent {\scriptsize \cite{zhang2018neural}} & 41.05 & 18.77 & 37.54 \\
Refresh {\scriptsize \cite{narayan-etal-2018-ranking}} & 41.00 & 18.80 & 37.70 \\
BanditSum {\scriptsize \cite{dong-etal-2018-banditsum}} & 41.50 & 18.70 & 37.60 \\
NeuSUM {\scriptsize \cite{zhou-etal-2018-neural}} & 41.59 & 19.01 & 37.98 \\
ExConSum {\scriptsize \cite{mendes-etal-2019-jointly}} & 41.70 & 18.60 & 37.80 \\
JECS {\scriptsize \cite{xu-durrett-2019-neural}} & 41.70 & 18.50 & 37.90 \\
LSTM+PN {\scriptsize \cite{zhong-etal-2019-closer}} & 41.85 & 18.93 & 38.13 \\ 
HER {\scriptsize \cite{luo-etal-2019-reading}} & 42.30 & 18.90 & 37.60  \\
HiBERT {\scriptsize \cite{zhang-etal-2019-hibert}} & 42.37 & 19.95 & 38.83 \\

PNBERT {\scriptsize \cite{zhong-etal-2019-searching}} & 42.69 & 19.60 & 38.85 \\

BERTSum {\scriptsize \cite{Liu2019TextSW}} & 42.61 & 19.99 & 39.09 \\
BERTSum+TriBlk & 43.25 & 20.24 & 39.63 \\

ARedSum-CTX {\scriptsize \cite{aredsum}} & 43.43 & 20.44 & \textbf{39.83} \\

HSG {\scriptsize \cite{wang-extsum-acl20}} & 42.31 & 19.51 & 38.74 \\
HSG+TriBlk & 42.95 & 19.76 & 39.23 \\

BERTSum Large* & 43.85 & 20.34 & 39.90 \\
\hline

\multicolumn{4}{c}{\textbf{Our non-stepwise models}} \\

\hline
BERTSum & 41.55 & 19.34 & 37.80\\
BERTSum+TriBlk & 42.70 & 19.93 & 38.89 \\
RoBERTaSum & 42.99 & 20.60 & 39.21 \\
RoBERTaSum+TriBlk & 43.30 & 20.58 & 39.48 \\


HiBERT & 41.43 & 19.23 & 37.73 \\
HiBERT+TriBlk  & 42.37 & 19.68 & 38.63 \\


ETCSum & 42.67 & 20.27 & 38.90 \\
ETCSum+TriBlk & 43.43 & 20.54 & 39.58 \\ \hline

\multicolumn{4}{c}{\textbf{Our stepwise models}} \\

\hline


Stepwise RoBERTaSum & 41.99 & 19.78 & 37.76\\
Stepwise RoBERTaSum+TriBlk & 41.50 & 19.48 & 37.25\\

Stepwise HiBERT & 41.98 & 19.53 & 38.32 \\
Stepwise HiBERT+TriBlk & 42.12 & 19.45 & 38.43 \\

Stepwise ETCSum & \textbf{43.84} & \textbf{20.80} & 39.77 \\

Stepwise ETCSum+TriBlk & 43.23 & 20.30 & 39.15 \\
\hline
\end{tabular}
\caption{Rouge F1 scores on the CNN/DailyMail test set. Boldfaced numbers are the best results among comparable models. * BERTSum Large builds on BERTLarge (24 layers) architectures, whereas ours build on BERTBase (12 layers) architectures.}
\label{tab:expcnndm}
\end{table}

We also trained non-stepwise variants of HiBERT and ETCSum models (the third block in Table~\ref{tab:expcnndm}). In this setting, HiBERT and ETC do not take partial summaries as input.
Instead, they simply take the input document and generate salient scores (using a sigmoid layer) for each sentence in the document; the top three sentences are then assembled to generate the summary. Our implementation of HiBERT differs from \newcite{zhang-etal-2019-hibert}. For example, we don't pretrain HiBERT from scratch for document modeling as in \newcite{zhang-etal-2019-hibert}. Instead, we initialize our HiBERT models with publicly available RoBERTa \cite{roberta} checkpoints following the superior performance of RoBERTaSum over BERTSum. We use different number of layers in the document encoder (L$_{\text{doc}}$ = 3) and in the sentence encoder (L$_{\text{sent}}$ = 9), as opposed to equal number of layers (L = 6) in both encoders of \newcite{zhang-etal-2019-hibert}. The layers in the document and sentence encoders were initialized with the top and the bottom layers of RoBERTa, respectively. All ETCSum models were initialized with the uncased version of ETC pretrained checkpoints \cite{etc} pretrained using the standard masked language model task and the contrastive predictive coding \cite{cpc}.\footnote{We thank the authors \cite{etc} for sharing their ETC checkpoints with us.}

We also report on the effect of TriBLK with all our models. We only experiment with the base-sized models and therefore have 12 layers, a hidden size of 768, filter size of 3072, and 12 attention heads. For comparison, we report results from BERTSum Large \cite{Liu2019TextSW} which uses 24 layers.
Finally, we employ a beam decoding to predict summaries using our stepwise models; we use a beam size of 3 for a maximum of 4 steps. We don't allow repeated sentences, though this is not a requirement. We refer the reader to the supplementary material for implementation and reproducibility details.


\pgfplotstableread[row sep=\\,col sep=&]{
length & Human & ETCSum & StepwiseETCSum & StepwiseETCSumTriBlk\\ 
1  &  0.0 & 0.0 & 0.0 & 0.0  \\ 
2  &  0.0 & 0.0 & 0.0 & 0.0  \\ 
3  &  0.0 & 0.0 & 0.0 & 0.0  \\ 
4  &  0.0 & 0.0 & 0.0 & 0.0  \\ 
5  &  0.0 & 0.0 & 0.0 & 0.0  \\ 
6  &  0.0 & 0.0 & 0.0 & 0.0  \\ 
7  &  0.0 & 0.0 & 0.0 & 0.0  \\ 
8  &  0.0 & 0.0 & 0.0 & 0.0  \\ 
9  &  8.735150244584206e-05 & 0.0 & 0.0 & 0.0  \\ 
10  &  0.0 & 0.0 & 0.0 & 0.0  \\ 
11  &  8.735150244584206e-05 & 0.0 & 0.0 & 0.0  \\ 
12  &  0.0 & 0.0 & 0.0 & 0.0  \\ 
13  &  0.0 & 0.0 & 0.0 & 0.0  \\ 
14  &  0.0002620545073375262 & 0.0 & 0.0 & 8.735150244584206e-05  \\ 
15  &  0.0002620545073375262 & 0.0 & 0.00017470300489168413 & 0.00043675751222921035  \\ 
16  &  0.0002620545073375262 & 0.0 & 0.00017470300489168413 & 0.0002620545073375262  \\ 
17  &  0.00034940600978336826 & 0.0 & 8.735150244584206e-05 & 0.00017470300489168413  \\ 
18  &  0.00043675751222921035 & 0.0 & 8.735150244584206e-05 & 0.0005241090146750524  \\ 
19  &  0.0008735150244584207 & 0.0 & 0.00017470300489168413 & 0.0007861635220125787  \\ 
20  &  0.0015723270440251573 & 0.0 & 0.0006114605171208945 & 0.001310272536687631  \\ 
21  &  0.0019217330538085255 & 0.0 & 0.0005241090146750524 & 0.0010482180293501049  \\ 
22  &  0.00253319357092942 & 0.0 & 0.00034940600978336826 & 0.0008735150244584207  \\ 
23  &  0.0032320055904961563 & 0.0 & 0.0007861635220125787 & 0.001222921034241789  \\ 
24  &  0.0033193570929419985 & 8.735150244584206e-05 & 0.001222921034241789 & 0.0014849755415793152  \\ 
25  &  0.004018169112508735 & 0.0 & 0.0009608665269042627 & 0.0014849755415793152  \\ 
26  &  0.004367575122292104 & 8.735150244584206e-05 & 0.0007861635220125787 & 0.0015723270440251573  \\ 
27  &  0.004891684136967156 & 0.0 & 0.00034940600978336826 & 0.0016596785464709992  \\ 
28  &  0.003668763102725367 & 0.0 & 0.0008735150244584207 & 0.0020964360587002098  \\ 
29  &  0.004979035639412998 & 0.0 & 0.0014849755415793152 & 0.00253319357092942  \\ 
30  &  0.004804332634521314 & 0.0 & 0.0017470300489168414 & 0.002183787561146052  \\ 
31  &  0.00550314465408805 & 8.735150244584206e-05 & 0.0020964360587002098 & 0.003843466107617051  \\ 
32  &  0.007337526205450734 & 0.0 & 0.0023584905660377358 & 0.0034940600978336828  \\ 
33  &  0.008385744234800839 & 0.0 & 0.002795248078266946 & 0.004367575122292104  \\ 
34  &  0.012491264849755416 & 8.735150244584206e-05 & 0.0034940600978336828 & 0.004979035639412998  \\ 
35  &  0.0159853249475891 & 0.0 & 0.0041928721174004195 & 0.0053284416491963665  \\ 
36  &  0.019304682040531096 & 0.00017470300489168413 & 0.004804332634521314 & 0.006464011180992313  \\ 
37  &  0.022187281621243886 & 0.00017470300489168413 & 0.006638714185883997 & 0.008123689727463312  \\ 
38  &  0.026991614255765198 & 0.0002620545073375262 & 0.006813417190775681 & 0.007774283717679944  \\ 
39  &  0.02620545073375262 & 0.00017470300489168413 & 0.005590496156533892 & 0.0069881201956673656  \\ 
40  &  0.023934311670160725 & 0.00043675751222921035 & 0.006464011180992313 & 0.008211041229909155  \\ 
41  &  0.021313766596785466 & 0.0006114605171208945 & 0.00759958071278826 & 0.009521313766596786  \\ 
42  &  0.018081761006289308 & 0.0005241090146750524 & 0.006638714185883997 & 0.00716282320055905  \\ 
43  &  0.019479385045422782 & 0.001222921034241789 & 0.006900768693221523 & 0.008123689727463312  \\ 
44  &  0.019042627533193572 & 0.001222921034241789 & 0.00803633822501747 & 0.009084556254367574  \\ 
45  &  0.015111809923130678 & 0.001222921034241789 & 0.009608665269042627 & 0.009958071278825996  \\ 
46  &  0.016596785464709992 & 0.0015723270440251573 & 0.008997204751921733 & 0.010220125786163521  \\ 
47  &  0.018780573025856045 & 0.0017470300489168414 & 0.007250174703004892 & 0.008735150244584208  \\ 
48  &  0.018343815513626835 & 0.001222921034241789 & 0.008123689727463312 & 0.009259259259259259  \\ 
49  &  0.023410202655485674 & 0.0017470300489168414 & 0.010918937805730259 & 0.0110062893081761  \\ 
50  &  0.022973445143256464 & 0.0030573025856044725 & 0.011443046820405312 & 0.012840670859538784  \\ 
51  &  0.02437106918238994 & 0.0034067085953878406 & 0.011355695317959469 & 0.012491264849755416  \\ 
52  &  0.026816911250873515 & 0.003668763102725367 & 0.00890985324947589 & 0.011443046820405312  \\ 
53  &  0.028825995807127882 & 0.004018169112508735 & 0.011879804332634521 & 0.012840670859538784  \\ 
54  &  0.0286512928022362 & 0.005153738644304682 & 0.011268343815513627 & 0.012054507337526206  \\ 
55  &  0.027341020265548566 & 0.00506638714185884 & 0.012928022361984625 & 0.014675052410901468  \\ 
56  &  0.026816911250873515 & 0.006726065688329839 & 0.016247379454926623 & 0.015548567435359888  \\ 
57  &  0.025069881201956672 & 0.008298392732354996 & 0.015111809923130678 & 0.015810621942697414  \\ 
58  &  0.01982879105520615 & 0.007686932215234102 & 0.01519916142557652 & 0.01642208245981831  \\ 
59  &  0.020003494060097833 & 0.009608665269042627 & 0.01685883997204752 & 0.019042627533193572  \\ 
60  &  0.018256464011180994 & 0.010744234800838574 & 0.0126659678546471 & 0.014150943396226415  \\ 
61  &  0.018693221523410204 & 0.011180992313067784 & 0.015024458420684835 & 0.01519916142557652  \\ 
62  &  0.01816911250873515 & 0.012054507337526206 & 0.017645003494060098 & 0.01781970649895178  \\ 
63  &  0.016771488469601678 & 0.011530398322851153 & 0.014937106918238994 & 0.01476240391334731  \\ 
64  &  0.015286512928022362 & 0.013190076869322153 & 0.014675052410901468 & 0.01476240391334731  \\ 
65  &  0.012753319357092943 & 0.013452131376659678 & 0.017208245981830888 & 0.01642208245981831  \\ 
66  &  0.012141858839972047 & 0.014587700908455625 & 0.01738294898672257 & 0.015810621942697414  \\ 
67  &  0.011879804332634521 & 0.01484975541579315 & 0.015461215932914047 & 0.01607267645003494  \\ 
68  &  0.011268343815513627 & 0.015461215932914047 & 0.01685883997204752 & 0.017470300489168415  \\ 
69  &  0.009346610761705102 & 0.017907058001397625 & 0.017120894479385047 & 0.016684136967155837  \\ 
70  &  0.009346610761705102 & 0.01650943396226415 & 0.016160027952480782 & 0.016160027952480782  \\ 
71  &  0.00890985324947589 & 0.015461215932914047 & 0.01484975541579315 & 0.01519916142557652  \\ 
72  &  0.008735150244584208 & 0.018256464011180994 & 0.015024458420684835 & 0.013976240391334731  \\ 
73  &  0.007948986722571627 & 0.018518518518518517 & 0.01519916142557652 & 0.01476240391334731  \\ 
74  &  0.00847309573724668 & 0.018693221523410204 & 0.01476240391334731 & 0.014675052410901468  \\ 
75  &  0.008822501747030049 & 0.01729559748427673 & 0.014412997903563941 & 0.015373864430468204  \\ 
76  &  0.007774283717679944 & 0.02122641509433962 & 0.012753319357092943 & 0.012578616352201259  \\ 
77  &  0.00803633822501747 & 0.02157582110412299 & 0.014937106918238994 & 0.014412997903563941  \\ 
78  &  0.006464011180992313 & 0.019916142557651992 & 0.014150943396226415 & 0.01310272536687631  \\ 
79  &  0.007861635220125786 & 0.020527603074772888 & 0.014063591893780572 & 0.013452131376659678  \\ 
80  &  0.006464011180992313 & 0.019566736547868623 & 0.016771488469601678 & 0.01519916142557652  \\ 
81  &  0.006464011180992313 & 0.022711390635918937 & 0.01519916142557652 & 0.014238294898672257  \\ 
82  &  0.004804332634521314 & 0.022187281621243886 & 0.015897973445143255 & 0.014587700908455625  \\ 
83  &  0.00550314465408805 & 0.022274633123689727 & 0.012403913347309574 & 0.012316561844863731  \\ 
84  &  0.0053284416491963665 & 0.021750524109014676 & 0.013888888888888888 & 0.014063591893780572  \\ 
85  &  0.005241090146750524 & 0.02157582110412299 & 0.012928022361984625 & 0.012491264849755416  \\ 
86  &  0.005677847658979734 & 0.02122641509433962 & 0.013277428371767994 & 0.013190076869322153  \\ 
87  &  0.004367575122292104 & 0.020440251572327043 & 0.013888888888888888 & 0.012928022361984625  \\ 
88  &  0.0053284416491963665 & 0.019916142557651992 & 0.012928022361984625 & 0.01310272536687631  \\ 
89  &  0.004542278127183787 & 0.019916142557651992 & 0.014412997903563941 & 0.012316561844863731  \\ 
90  &  0.004280223619846262 & 0.01939203354297694 & 0.011617749825296996 & 0.0110062893081761  \\ 
91  &  0.003756114605171209 & 0.018256464011180994 & 0.012491264849755416 & 0.01179245283018868  \\ 
92  &  0.003756114605171209 & 0.020440251572327043 & 0.011355695317959469 & 0.011180992313067784  \\ 
93  &  0.004105520614954577 & 0.015286512928022362 & 0.011093640810621943 & 0.00969601677148847  \\ 
94  &  0.0030573025856044725 & 0.019566736547868623 & 0.012141858839972047 & 0.01056953179594689  \\ 
95  &  0.0032320055904961563 & 0.01642208245981831 & 0.011617749825296996 & 0.009958071278825996  \\ 
96  &  0.0030573025856044725 & 0.015810621942697414 & 0.010482180293501049 & 0.009958071278825996  \\ 
97  &  0.002795248078266946 & 0.015810621942697414 & 0.011355695317959469 & 0.009084556254367574  \\ 
98  &  0.0022711390635918936 & 0.014238294898672257 & 0.011443046820405312 & 0.009608665269042627  \\ 
99  &  0.002707896575821104 & 0.015286512928022362 & 0.010045422781271837 & 0.008647798742138365  \\ 
100  &  0.00253319357092942 & 0.013626834381551363 & 0.010220125786163521 & 0.009171907756813417  \\ 
101  &  0.0022711390635918936 & 0.012928022361984625 & 0.007774283717679944 & 0.006027253668763103  \\ 
102  &  0.0032320055904961563 & 0.0110062893081761 & 0.008298392732354996 & 0.007948986722571627  \\ 
103  &  0.00253319357092942 & 0.0126659678546471 & 0.007948986722571627 & 0.008647798742138365  \\ 
104  &  0.0014849755415793152 & 0.012316561844863731 & 0.006726065688329839 & 0.006813417190775681  \\ 
105  &  0.00253319357092942 & 0.01179245283018868 & 0.006900768693221523 & 0.006900768693221523  \\ 
106  &  0.0017470300489168414 & 0.009870719776380153 & 0.006289308176100629 & 0.005153738644304682  \\ 
107  &  0.0020090845562543676 & 0.010220125786163521 & 0.006813417190775681 & 0.005415793151642208  \\ 
108  &  0.0020090845562543676 & 0.007075471698113208 & 0.0058525506638714185 & 0.006114605171208945  \\ 
109  &  0.0020964360587002098 & 0.00969601677148847 & 0.0069881201956673656 & 0.006201956673654787  \\ 
110  &  0.0017470300489168414 & 0.009958071278825996 & 0.006638714185883997 & 0.0053284416491963665  \\ 
111  &  0.001397624039133473 & 0.007774283717679944 & 0.005415793151642208 & 0.005153738644304682  \\ 
112  &  0.0014849755415793152 & 0.006813417190775681 & 0.0053284416491963665 & 0.00506638714185884  \\ 
113  &  0.0014849755415793152 & 0.0058525506638714185 & 0.0047169811320754715 & 0.004629629629629629  \\ 
114  &  0.0020090845562543676 & 0.007424877707896575 & 0.00550314465408805 & 0.004629629629629629  \\ 
115  &  0.001222921034241789 & 0.007075471698113208 & 0.005241090146750524 & 0.0047169811320754715  \\ 
116  &  0.0010482180293501049 & 0.005590496156533892 & 0.00550314465408805 & 0.0041928721174004195  \\ 
117  &  0.001222921034241789 & 0.0047169811320754715 & 0.004105520614954577 & 0.003843466107617051  \\ 
118  &  0.0008735150244584207 & 0.005241090146750524 & 0.003756114605171209 & 0.002620545073375262  \\ 
119  &  0.0006988120195667365 & 0.0047169811320754715 & 0.0034067085953878406 & 0.0029699510831586303  \\ 
120  &  0.0007861635220125787 & 0.004280223619846262 & 0.003581411600279525 & 0.0034940600978336828  \\ 
121  &  0.0006114605171208945 & 0.004105520614954577 & 0.0034940600978336828 & 0.0032320055904961563  \\ 
122  &  0.0015723270440251573 & 0.004018169112508735 & 0.002445842068483578 & 0.0022711390635918936  \\ 
123  &  0.0006988120195667365 & 0.003581411600279525 & 0.0031446540880503146 & 0.0030573025856044725  \\ 
124  &  0.0007861635220125787 & 0.002795248078266946 & 0.00253319357092942 & 0.0018343815513626835  \\ 
125  &  0.0007861635220125787 & 0.002882599580712788 & 0.002183787561146052 & 0.002620545073375262  \\ 
126  &  0.0002620545073375262 & 0.002183787561146052 & 0.0029699510831586303 & 0.0020090845562543676  \\ 
127  &  0.0006988120195667365 & 0.002795248078266946 & 0.0029699510831586303 & 0.002445842068483578  \\ 
128  &  0.00034940600978336826 & 0.0017470300489168414 & 0.0018343815513626835 & 0.001397624039133473  \\ 
129  &  0.0005241090146750524 & 0.002183787561146052 & 0.0019217330538085255 & 0.001397624039133473  \\ 
130  &  0.0007861635220125787 & 0.0017470300489168414 & 0.0023584905660377358 & 0.0020090845562543676  \\ 
131  &  0.00034940600978336826 & 0.0018343815513626835 & 0.0023584905660377358 & 0.0015723270440251573  \\ 
132  &  0.0006988120195667365 & 0.0014849755415793152 & 0.0018343815513626835 & 0.001310272536687631  \\ 
133  &  0.0006988120195667365 & 0.001222921034241789 & 0.0015723270440251573 & 0.001310272536687631  \\ 
134  &  0.00034940600978336826 & 0.0017470300489168414 & 0.0014849755415793152 & 0.0011355695317959468  \\ 
135  &  0.0002620545073375262 & 0.0007861635220125787 & 0.0017470300489168414 & 0.001397624039133473  \\ 
136  &  0.0008735150244584207 & 0.0011355695317959468 & 0.0010482180293501049 & 0.0007861635220125787  \\ 
137  &  0.0005241090146750524 & 0.0007861635220125787 & 0.0015723270440251573 & 0.0008735150244584207  \\ 
138  &  0.00034940600978336826 & 0.0005241090146750524 & 0.0011355695317959468 & 0.0002620545073375262  \\ 
139  &  0.00017470300489168413 & 0.0007861635220125787 & 0.0008735150244584207 & 0.0010482180293501049  \\ 
140  &  0.00043675751222921035 & 0.0005241090146750524 & 0.0011355695317959468 & 0.0007861635220125787  \\ 
141  &  0.0005241090146750524 & 0.0009608665269042627 & 0.0009608665269042627 & 0.00034940600978336826  \\ 
142  &  0.0002620545073375262 & 0.00017470300489168413 & 0.0010482180293501049 & 0.0006114605171208945  \\ 
143  &  0.00017470300489168413 & 0.00043675751222921035 & 0.0006114605171208945 & 0.00043675751222921035  \\ 
144  &  0.00034940600978336826 & 0.0005241090146750524 & 0.001222921034241789 & 0.0006988120195667365  \\ 
145  &  0.00017470300489168413 & 8.735150244584206e-05 & 0.0007861635220125787 & 0.00034940600978336826  \\ 
146  &  0.0002620545073375262 & 0.0002620545073375262 & 0.00043675751222921035 & 0.0008735150244584207  \\ 
147  &  0.00043675751222921035 & 0.00043675751222921035 & 0.0005241090146750524 & 0.00034940600978336826  \\ 
148  &  0.0 & 0.0002620545073375262 & 0.00034940600978336826 & 0.0002620545073375262  \\ 
149  &  0.00034940600978336826 & 0.0002620545073375262 & 0.00043675751222921035 & 0.00043675751222921035  \\ 
150  &  0.0 & 8.735150244584206e-05 & 0.00043675751222921035 & 0.00043675751222921035  \\ 
151  &  0.00017470300489168413 & 0.0002620545073375262 & 0.0005241090146750524 & 0.0002620545073375262  \\ 
152  &  0.00017470300489168413 & 0.0002620545073375262 & 0.00034940600978336826 & 0.00017470300489168413  \\ 
153  &  0.00034940600978336826 & 0.0002620545073375262 & 0.00043675751222921035 & 0.0002620545073375262  \\ 
154  &  0.0002620545073375262 & 0.00017470300489168413 & 0.0002620545073375262 & 0.00017470300489168413  \\ 
155  &  0.00017470300489168413 & 0.00017470300489168413 & 0.0 & 0.0  \\ 
156  &  0.0002620545073375262 & 8.735150244584206e-05 & 8.735150244584206e-05 & 0.00017470300489168413  \\ 
157  &  0.00017470300489168413 & 8.735150244584206e-05 & 0.0002620545073375262 & 0.00017470300489168413  \\ 
158  &  0.00017470300489168413 & 0.0002620545073375262 & 0.0002620545073375262 & 8.735150244584206e-05  \\ 
159  &  8.735150244584206e-05 & 0.0002620545073375262 & 8.735150244584206e-05 & 0.00017470300489168413  \\ 
160  &  8.735150244584206e-05 & 0.0 & 0.00034940600978336826 & 0.00043675751222921035  \\ 
161  &  8.735150244584206e-05 & 0.00017470300489168413 & 0.00017470300489168413 & 8.735150244584206e-05  \\ 
162  &  0.0 & 0.00017470300489168413 & 0.00017470300489168413 & 8.735150244584206e-05  \\ 
163  &  0.0 & 0.0 & 0.00017470300489168413 & 0.00017470300489168413  \\ 
164  &  0.00017470300489168413 & 0.0 & 0.0 & 0.0  \\ 
165  &  0.0 & 0.0 & 0.0 & 8.735150244584206e-05  \\ 
166  &  0.0 & 8.735150244584206e-05 & 0.0002620545073375262 & 0.0  \\ 
167  &  8.735150244584206e-05 & 8.735150244584206e-05 & 8.735150244584206e-05 & 8.735150244584206e-05  \\ 
168  &  8.735150244584206e-05 & 0.0 & 0.0 & 0.0  \\ 
169  &  0.0 & 0.00017470300489168413 & 0.00017470300489168413 & 0.0  \\ 
170  &  8.735150244584206e-05 & 0.0 & 8.735150244584206e-05 & 8.735150244584206e-05  \\ 
171  &  8.735150244584206e-05 & 0.0 & 8.735150244584206e-05 & 0.0  \\ 
172  &  8.735150244584206e-05 & 0.0 & 0.0 & 0.0  \\ 
173  &  8.735150244584206e-05 & 0.0 & 8.735150244584206e-05 & 8.735150244584206e-05  \\ 
174  &  0.0 & 8.735150244584206e-05 & 0.0 & 0.0  \\ 
175  &  0.0 & 0.0 & 0.0 & 0.0  \\ 
176  &  8.735150244584206e-05 & 0.0 & 0.0 & 0.0  \\ 
177  &  0.0 & 0.0 & 0.0 & 0.0  \\ 
178  &  8.735150244584206e-05 & 0.0 & 0.0 & 8.735150244584206e-05  \\ 
179  &  0.0 & 0.0 & 0.0 & 0.0  \\ 
180  &  0.0 & 0.0 & 0.0 & 0.0  \\ 
181  &  0.00017470300489168413 & 0.0 & 0.0 & 0.0  \\ 
182  &  0.0 & 0.0 & 0.0 & 0.0  \\ 
183  &  0.0 & 0.0 & 0.0 & 0.0  \\ 
184  &  0.0 & 0.0 & 0.0 & 0.0  \\ 
185  &  8.735150244584206e-05 & 0.0 & 0.0 & 0.0  \\ 
186  &  0.0 & 0.0 & 0.0 & 0.0  \\ 
187  &  0.0 & 0.0 & 0.0 & 0.0  \\ 
188  &  8.735150244584206e-05 & 0.0 & 0.0 & 0.0  \\ 
189  &  8.735150244584206e-05 & 0.0 & 0.0 & 0.0  \\ 
190  &  0.0 & 0.0 & 0.0 & 0.0  \\ 
191  &  0.0 & 0.0 & 0.0 & 0.0  \\ 
192  &  0.00017470300489168413 & 0.0 & 0.0 & 0.0  \\ 
193  &  0.0 & 0.0 & 8.735150244584206e-05 & 0.0  \\ 
194  &  0.0 & 0.0 & 0.0 & 0.0  \\ 
195  &  0.0 & 0.0 & 0.0 & 0.0  \\ 
196  &  0.0 & 0.0 & 0.0 & 0.0  \\ 
197  &  0.0 & 0.0 & 0.0 & 0.0  \\ 
198  &  8.735150244584206e-05 & 0.0 & 0.0 & 0.0  \\ 
199  &  0.0 & 0.0 & 8.735150244584206e-05 & 0.0  \\ 
200  &  0.0 & 0.0 & 0.0 & 0.0  \\ 
}\rougetwo

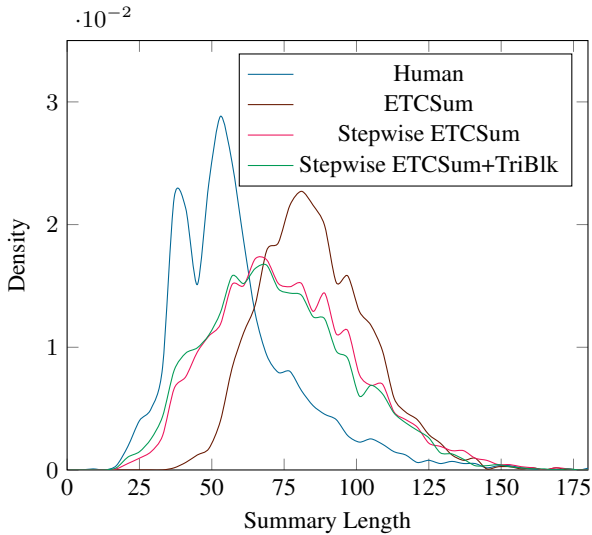
\begin{figure}[t!]
  \footnotesize
  \center{
    \begin{tikzpicture}
      \begin{axis}[
        title={},
        xlabel={Summary Length},
        ylabel={Density},
        xmin=0, xmax=180,
        ymin=0, ymax=0.035,
        xtick={0, 25, 50, 75, 100, 125, 150, 175},
        xticklabels={0, 25, 50, 75, 100, 125, 150, 175},
        legend pos=north east,
        samples=500,
        ]
        \addplot [midnightblue,smooth,each nth point=4] table[x=length,y=Human]{\rougetwo};
        \addplot [sepia,smooth,each nth point=4] table[x=length,y=ETCSum]{\rougetwo};
        \addplot [orangered,smooth,each nth point=4] table[x=length,y=StepwiseETCSum]{\rougetwo};
        \addplot [forestgreen,smooth,each nth point=4] table[x=length,y=StepwiseETCSumTriBlk]{\rougetwo};
        \legend{Human, ETCSum, Stepwise ETCSum, Stepwise ETCSum+TriBlk}
      \end{axis}
    \end{tikzpicture}
  }
  \vspace{-0.5cm}
  \caption{Length distributions in ETCSum summaries on the CNN/DailyMail test set.\label{fig:etcsum-length}}
  \vspace{-0.6cm}
\end{figure}

\paragraph{Generating Extractive Oracles.}
Following \newcite{narayan-etal-2018-ranking}, we train models to predict all sentences in Oracle (Full) for non-stepwise training. Stepwise training learns to do this gradually: at each step, we train model to predict the next sentence in Oracle (Full) using the earlier predicted sentences and the document. During testing, human written abstracts are used as reference summaries to evaluate our models. 

\begin{table*}[t!]
\centering
\footnotesize
\begin{threeparttable}
\begin{tabular}{l|c|ccc|c|c}
\hline
\multirow{2}{*}{Models} & \multicolumn{1}{c|}{RG} & \multicolumn{3}{c|}{CS} & CO & \multirow{2}{*}{BLEU} \\
& P\% & P\% & R\% & F1\% & DLD\% & \\ \hline
CC \cite{wiseman-etal-2017-challenges} & 74.80 & 29.49 & 36.18 & 32.49 & 15.42 & 14.19 \\
NCP+CC \cite{rotowire-ncpcc} & 87.47 & 34.18 & 51.22 & 41.00 & 18.58 & 16.50 \\
HierEnc \cite{gong-etal-2019-table} & 91.46 & 36.09 & 48.01 & 41.21 & 20.86 & 16.85 \\
EdiNLG \cite{rotowire-edinlg} & 91.41 & 30.91 & \textbf{64.13} & 41.71 & 21.72 & 17.01 \\
MS-GPT-50 \cite{rotowire-msgpt} & 94.35 & 33.91 & 53.82 & 41.61 & 19.30 & 15.17 \\
MS-End-to-End \cite{rotowire-msgpt} & 93.38 & 32.40 & 58.02 & 41.58 & 18.54 & 15.03 \\
Systran-AI-Detok \cite{rotowire-systran} & 84.16 & 34.88 & 43.29 & 38.63 & 22.72 & \textbf{18.32} \\
NLE* \cite{rotowire-nle} & 94.08 & 41.13 & 54.20 & 46.77 & 25.64 & 20.52 \\
Hierarchical D2T \cite{rotowire-hierarchical} & 89.46 & 39.47 & 51.64 & 44.74 & 18.90 & 17.50 \\
\hline
Stepwise HiBERT realized & 95.88 & 41.49 & 53.86 & 46.87 & 18.10 & 14.79 \\ 
Stepwise HiBERT planning only* & -- & 42.96 & 55.81 & 48.55 & -- & -- \\ 
Stepwise ETCSum realized & \textbf{98.87} & \textbf{45.79} & 58.49 & \textbf{49.76} & \textbf{25.08} & 17.56 \\
Stepwise ETCSum planning only* & -- & 46.02 & 58.45 & 51.50 & -- & -- \\
\hline
\end{tabular}

\end{threeparttable}
\vspace{-0.1cm}
\caption{Standard metrics for Rotowire: relation generation (RG) precision (P\%), content selection (CS) precision (P\%) and recall (R\%), content ordering (CO) via the complement of normalized Damerau-Levenshtein distance (DLD\%), and BLEU score. Models marked with a * are not directly comparable. Boldfaced numbers are the best results among comparable models.}
\label{table:rotowire-results}
\vspace{-0.5cm}
\end{table*}

\subsection{Results}

\paragraph{Long form Summarization.} 

In our experiments, ETCSum appears to be far more superior than HiBERT when modeling long documents for extractive summarization; ETCSum outperformed HiBERT in all cases including stepwise or non-stepwise predictions, and, with or without trigram blocking.  
The downside of HiBERT where token-level attention across sentences is not possible, is not optimal for modeling documents. Both ETCSum and ETCSum+TriBlk performed better than BERTSum and BERTSum+TriBlk, respectively. These results suggest the importance of modeling the whole document with ETCSum, rather than truncating it to only 512 tokens to fit BERTSum. However, the improvement may not be attributed solely to ETCSum's ability to model long inputs, but also to its better initialization with ETC checkpoints \cite{etc}, specially when the improvement diminishes when compared against RoBERTaSum.\footnote{One may consider to access the modeling of long inputs in ETCSum against the truncated inputs in BERTSum and RoBERTaSum, by initializing ETCSum with BERT or RoBERTa checkpoints, and not ETC checkpoint. However, this is not fair to ETCSum as BERT or RoBERTa uses absolute position embeddings \cite{bert}, whereas, ETC uses relative position embeddings \cite{shaw-etal-2018-self}.}

\paragraph{Stepwise vs Non-stepwise models.}

First of all, trigram filtering seems to be the key in addressing redundancy in generated summaries in non-stepwise models. It helps almost all models including our HiBERT and ETCSum (except for the single case of RoBERTaSum on Rouge-2). Interestingly, we don't observe the same pattern for our stepwise models. We observe that our stepwise models (both HiBERT and ETCSum, without TriBlk) consistently improve over their non-stepwise counterparts. But when stepwise is applied with TriBlk, we don't always see improvements. 
We conjecture that our stepwise models themselves are inherently better at avoiding redundancy in generated summaries due to the knowledge of previously generated summary at each prediction step, and improvements with TriBlk are not always complementary. The same is also demonstrated in Figure~\ref{fig:etcsum-length}; density curves show that Stepwise ETCSum (avg:76.96, std:24.77) follows the human distribution (avg:58.3, std:24.8) better than ETCSum (avg:85.84, std:19.06). With Stepwise ETCSum+TriBlk (avg:73.92, std:24.76), we don't see significant improvement over Stepwise ETCSum.

We also report on Stepwise RoBERTaSum baselines and performance dropped compared to corresponding non-stepwise models. Perhaps without any structure in the transformer, simple summary concatenation is not a good method for Stepwise RoBERTaSum to distinguish the document from the summary. There might be better ways (than the vanilla concatenation), but with Stepwise ETCSum or HiBERT, it is very natural. Stepwise RoBERTaSum also loses access to the end of the input as the partial summary grows for documents that are already close to 512 tokens in length.

Finally, our Stepwise ETCSum model without any explicit redundancy or sentence selection mechanisms, achieved comparable performance to the state of the art on the CNN/DailyMail extractive summarization task with a smaller model; BERTSum Large \cite{Liu2019TextSW} with 340m parameters achieved 43.85/20.34/39.90 R1/R2/RL scores, whereas ours with 165m parameters achieved 43.84/20.80/39.77. Comparatively, Stepwise HiBERT did not do equally well on document summarization due to the sequential nature of the input. However, we demonstrate in Section~\ref{sec:exprotowire} that it is well suited as an extractive content planner for table-to-text generation. 

ROUGE scores in Table~\ref{tab:expcnndm} are computed with a confidence interval of 95\%. As such, Stepwise ETCSum(+TriBlk) is significantly better than BERTSum(+TriBlk), all variants of HierBERT, ETCSum and Stepwise RoBERTaSum(+TriBlk). For other models, such as RoBERTaSum(+TriBlk) and ETCSum+TriBlk, this confidence interval is not a deciding factor, hence we performed One-way ANOVA with posthoc Tukey-HSD tests ($p<0.01$). Our best model Stepwise ETCSum performs significantly better than RoBERTaSum(+TriBlk), ETCSum+TriBlk and Stepwise ETCSum+TriBlk, on the average of ROUGE scores.

\section{Table-to-Text Generation}
\label{sec:exprotowire}

\paragraph{Task.}

We further explore our model's ability to learn content plans for the Rotowire data-to-text generation task \cite{wiseman-etal-2017-challenges}.\footnote{The Rotowire dataset is available for download at \url{https://github.com/harvardnlp/boxscore-data}.} The task is to generate a summary of an NBA game from its box score (a table of statistics detailing the performance of the two teams and of each player). 
The dataset consists of 4853 pairs of box scores and summaries of NBA games played from 2014 to 2017. 
The data is split into 3398 train, 727 validation and 728 test examples. On average there are 628 records in a box score per game. The average summary has 337.1 words and 13.49 sentences.

Similar to \newcite{rotowire-ncpcc} we decompose the problem into two sub-problems, which we solve independently: content planning, which consists of selecting which records in the table should be mentioned in the summary, in what order, and how they should be organized into sentences; and realization, which uses the content plan to create a human-readable summary. We refer the reader to the supplementary material for an example.
Our main focus in this paper is to demonstrate our models' ability to model long and structured Rotowire input tables, and generate long meaningful content plans. For realization, we simply use a RoBERTa \cite{roberta} initialized sequence-to-sequence transformer model \cite{berts2s}, trained to emit the realization sentence by sentence. 

We train our stepwise models to take a score table and the partially generated content plan, and predict the next element in the content plan. This can be either one of the entries in the score table, a sentence break or a token marking the end of the plan. 
Unlike extractive summarization, here an optimal extractive content plan can have repeated entries from the input table (e.g. team names) to better preserve and generate discourse relations among sentences in the target summary \cite{puduppully-etal-2019-data}, making it a challenging task for other iterative models that prohibit redundancy, e.g., \cite{aredsum}. 
For details about model implementation, realization, and the induction of oracle content plans for training, we refer the reader to the supplementary material.

We report typical Rotowire metrics \cite{wiseman-etal-2017-challenges}, using the standard information extraction system described by \newcite{rotowire-ncpcc} to extract the box score table relations mentioned in the generated (G) and in the target (T) summary. The metrics measure: text quality (BLEU score between G and T); relation generation quality (the precision of the relations extracted from G against the box score table); content selection quality (the precision and recall of the relations extracted from G against those extracted from T); and content ordering quality (the complement of the normalized Damerau-Levenshtein distance on the sequences of relations extracted from G and T). We also conducted human evaluation of Rotowire summaries. 

\paragraph{Results.} 

We focus on evaluating our Stepwise HiBERT and ETCSum models.\footnote{We don't reproduce BERTSum or RoBERTaSum baselines here for two reasons: i) these sequential models are not optimal for tabular data, and ii) they are also bounded by an input length of 512 tokens, the average length of linearized score tables is 7184 tokens per game. We also don't report on our non-stepwise models as they are not suitable to generate ordered content plans as required for this task.} Our results are presented in Table~\ref{table:rotowire-results}. The ``realized'' scores assess the quality of our realized summaries and are comparable to systems in the first block in Table~\ref{table:rotowire-results}. We found both Stepwise HiBERT and Stepwise ETCSum do content selection particularly well. Their very high precision scores (41.49\% and 45.79\%, respectively) combined with good recall (53.86\% and 58.49\%, respectively) outperform \newcite{rotowire-ncpcc} and other recent models on F1 score.
In terms of content ordering and BLEU score, Stepwise HiBERT (14.79 BLEU, 18.10\% DLD) performs worse than \newcite{rotowire-ncpcc} (16.50 BLEU, 18.58\% DLD), while Stepwise ETCSum performs significantly better (17.56 BLEU, 25.08\% DLD).
It's possible that a higher BLEU score could be achieved by improving our simple sentence-by-sentence realization method.

We also report content selection scores for the output of the content planning modules (see ``planning only'' models in Table~\ref{table:rotowire-results}). We drop name, city and date entries from our content plans before computing the metrics in order to make them comparable with others in Table~\ref{table:rotowire-results}. We see the roundtrip of realization and subsequent information extraction decreases CS quality slightly for both models (the absolute drop of F1 score is 1.68\% for Stepwise HiBERT, and 1.74\% for Stepwise ETCSum).

\begin{table}[t!]
\centering
\footnotesize
\begin{tabular}{l|cc}
\hline
Models & Informativeness & Readability \\ \hline
Baseline & 0.06 & \textbf{0.22} \\
Stepwise HiBERT & 0.17 & 0.00 \\
\hspace{0.5cm} +truncated & -0.34 & 0.04 \\
Stepwise ETCSum & \textbf{0.29} & -0.13 \\
\hspace{0.5cm} +truncated & -0.10 & -0.08\\
Gold & -0.09 & -0.03 \\
\hline
\end{tabular}
\vspace{-0.1cm}
\caption{Human evaluation of Rotowire Summaries.}
\label{table:rotowire-humaneval}
\vspace{-0.5cm}
\end{table}



\paragraph{Human Evaluation.}  

Participants were shown two summaries of an NBA game and asked to compare them with respect to {\em informativeness} (Does a summary present a better selection of the relevant facts about the game?) and {\em readability} (Which summary has a better narrative flow and is easier to read?). We randomly selected 50 NBA tables and evaluated summaries from Baseline  \cite{wiseman-etal-2017-challenges}, Stepwise HiBERT, Stepwise ETC and Gold. The average(max;min) number of sentences were 8(8;8), 12.7(17;9), 16.7(25;10) and 12.0(20;6), for Baseline, Stepwise HiBERT, Stepwise ETC, and Gold, respectively. We also included truncated summaries from Stepwise HiBERT and Stepwise ETC to match the number of sentences in corresponding Gold summaries. We elicited judgements from three different annotators for each pair. We report the Best(1)-Worst(-1) Scaling scores \cite{louviere1991best,louviere2015best}. Results are presented in Table~\ref{table:rotowire-humaneval}.

Overall, Stepwise ETC summaries were ranked most informative, but they performed worst on readability. The off-the-shelf sentence-level realizer (see the supplementary material) favors the statistics-dense sentences of the baseline summaries, as it tends to hallucinate on less dense plans. Future work will aim to address this limitation.
For infromativeness, Stepwise ETC summaries are significantly better than Gold, Stepwise ETC truncated and Stepwise HiBERT truncated summaries. Stepwise HiBERT summaries are significantly better than both truncated variants. All other differences are not significant ($p<0.05$). For readability, baseline summaries are significantly better than both ETC variants and Stepwise HiBERT. All other differences are not significant.

\section{Conclusion}

The stepwise structured transformer paradigm, exemplified by HiBERT and ETCSum, can be easily adapted both to extractive document summarization or content planning for table-to-text generation. 
Stepwise ETCSum, in particular, sets a new standard for both tasks. Future work will focus on extending our models to generate extractive plans for better abstractive summarization of long or multiple documents \cite{wikisum}.

\section*{Acknowledgments}

We thank Joshua Ainslie and Santiago Ontanon for sharing their ETC code and checkpoints, and also giving us feedback on an early draft of this paper.
We thank Annie Louis, the London and Zurich Generation teams, the reviewers and the action editor for invaluable feedback. We thank Enrique Alfonseca and Hadar Shemtov for their support for Longform Summarization. 

\bibliography{extsum} 
\bibliographystyle{acl_natbib}

\appendix

\section{Implementation and Reproducibility details}

\subsection{HiBERT} 

We did a wide range of hyperparameter search for HiBERT. We experimented with the number of layers in the document encoder ($1< L_{\text{doc}} < 12$); the number of layers in the sentence encoder ($1 < L_{\text{sent}} < 12, L_{\text{doc}} < L_{\text{sent}}$); the initialization and sharing of position embeddings, $\mathbf{p}^{\text{token}}_j$, $\mathbf{p}^{\text{doc}}_j$ and $\mathbf{p}^{\text{sum}}_j$; the initialization and sharing of document and sentence encoder parameters with BERT and RoBERTa checkpoints; and the representation of sentence (``first token embedding'' or ``average of all token embeddings'') from the sentence encoder.

For extractive summarization, we used HiBERT with a 8 transformer layer sentence encoder, and a 4 transformer layer document encoder. The model has 133,784,833 parameters. The word position embedding in the sentence encoder is initialized using the RoBERTa checkpoint, but the document and summary sentence position embeddings are learned from scratch. The document self attention and summary self attentions are shared and initialized using the RoBERTa checkpoint, the document-summary attention is also initialized using the RoBERTa checkpoint. We truncate each document to 128 sentences and each sentence to 32 words. We trained all HiBERT models for 100k steps saving checkpoints every 1000 steps, with a batch size of 32. Following \newcite{Liu2019TextSW}, we choose the best model based on the MLE loss on the whole validation set.

For Rotowire, we use HiBERT with a 2 transformer layer sentence encoder, and a 4 transformer layer document encoder. The model has 91,448,065 trainable parameters. We don't use the document sentence position embeddings for Rotowire as the input consists of a set of entries in a table. We use the summary sentence position embedding to capture the order in the content plan. We use the ROBERTA vocabulary, but as discussed in \ref{planning-details} we don't use ROBERTA pretraining, instead initializing with random weights.
We trained the model with a batch size of 128 until the AUC score for predicting the next content plan entry on the validation dataset flattened out, which came after 766K steps.
Since the dataset has 246290 examples (one for each element in the target content plan for each Rotowire example), the model saw the entire dataset approximately 398 times.

For all HiBERT models, we used Cloud TPU v3 accelerators for training and the Adam optimizer with a learning rate of 0.01.

\subsection{ETCSum} 

The ETCSum model for both extractive summarization and table-to-text generation uses a 12 layer transformer as described in \cite{etc}. The model is pretrained with MLM and CPC objectives as described in \cite{etc}. In total, the model has 165,825,793 trainable parameters which mostly comes from the long input of 8192 tokens and the full attention of 512 of the global tokens. We trained our model with a batch size of 512 for 5,000 steps approximately equivalent to 10 epochs.

We used Cloud TPU v3 accelerators for training and inference was done on a V100GPU taking 10 hours to get predictions for the test set.

Model selection was done over models Rouge-1 performance in the validation set for all models except stepwise models where a subset of the validation set was used instead, consisting of the first 1000 examples, given the longer inference times.

We did a wide range of hyperparameter search where we experimented with learning rate (0.000025, 0.00005, 0.0001), relative position encoding vocabulary size (12, 24), the representation of sentences (``first token embedding'' or ``average of all token embeddings'') from the sentence encoder and in additionally non-stepwise models we experimented with  positive label weight used to for loss calculation.
Finally, we used an Adam optimizer with  learning rate of 0.000025.

\subsection{Realization model} 

We use a ROBERTASHARE model following \cite{berts2s}. The model has 152,491,008 trainable parameters. We trained the model until we reached the maximum BLEU score on validation data. We trained our model with a batch size of 512 for 36K steps.
Since the dataset has 45533 examples (one for each element in the target content plan in each Rotowire example), the model saw the entire dataset approximately 405 times. We used Cloud TPU v3 accelerators for training. We used the Adam optimizer with a learning rate of 0.05.

\section{Table-to-Text Generation}

\begin{table*}[t!]
\centering
\footnotesize
\begin{tabular}{l|cccccccc}
\hline
\multicolumn{9}{c}{\textbf{Input Table:} Match date: Saturday, 22nd October 2018} \\
\hline
\textbf{Team} & Name & City & At home? & Wins & Losses & Points & Rebounds & ... \\
\hline
Chicago\_Bulls & Bulls & Chicago & Home & 3 & 1 & 100 & 21 & ... \\
LA\_Lakers & Lakers & Los\_Angeles & Away & 2 & 5 & 80 & 25 & ... \\
\hline
\textbf{Player} & Name & Surname & Team & Points & Rebounds & Assists & ... \\
\hline
Michael\_Jordan & Michael & Jordan & Chicago\_Bulls & 25 & 10 & 10 & ... \\
Shaquille\_O\_Neal & Shaquille & O'Neal & LA\_Lakers & 30 & 15 & 11 & ... \\
... & ... & ... & ... & ... & ... & ... & ... \\
\hline
\multicolumn{9}{c}{\textbf{Content Plan}} \\
\hline
\multicolumn{9}{p{15cm}}{\textbf{S1}: Chicago\_Bulls city, Chicago\_Bulls name, LA\_Lakers city, LA\_Lakers name, Chicago\_Bulls points, LA\_Lakers points, Match date, EOS.} \\
\multicolumn{9}{p{15cm}}{\textbf{S2}: LA\_Lakers name, LA\_Lakers wins, LA\_Lakers losses, Shaquille\_O\_Neal surname,  Shaquille\_O\_Neal points, EOS.} \\
\multicolumn{9}{p{15cm}}{\textbf{S3}: Chicago\_Bulls city, Chicago\_Bulls name, Chicago\_Bulls wins, Chicago\_Bulls losses, Michael\_Jordan name, Michael\_Jordan surname, Michael\_Jordan points, Michael\_Jordan rebounds, EOS.} \\
\hline
\multicolumn{9}{c}{\textbf{Realization}} \\
\hline
\multicolumn{9}{p{15cm}}{\textbf{S1}: The Chicago Bulls won against the Los Angeles Lakers 100 - 80 on Saturday.} \\
\multicolumn{9}{p{15cm}}{\textbf{S2}: It was a poor showing for the Lakers (2 - 5) in spite of O'Neal's 30 point contribution.} \\
\multicolumn{9}{p{15cm}}{\textbf{S3}: The Chicago Bulls' (3 - 1) best player was, predictably, Michael Jordan with 25 points and 10 rebounds. } \\
\hline
\end{tabular}
\caption{An hypothetical example from the Rotowire dataset for an NBA game, possible 3-sentence content plan and corresponding 3 realized sentences below. 
}
\label{table:boxscore}
\end{table*}

\subsection{Task}

Table~\ref{table:boxscore} shows a prototypical input table from the Rotowire dataset\footnote{We are not presenting an actual example for legal reasons.}, along with a possible content plan and its realization. As shown in the example, a well-formed content plan can repeat some of the entries from the input table. 

\subsection{Generating Oracle Content Plans} 
The Rotowire dataset does not contain ground truth content plans for its summaries. Instead, we infer them following a similar approach to \cite{rotowire-ncpcc}, but with a few minor modifications: 1) we use just a single convolutional model, instead of an ensemble of convolutional models and LSTMs, 2) our plans maintain the within-sentence order of information, and may include repetitions if a piece of information is repeated within a sentence in the target summary, 3) our plans include sentence breaks, though we remove sentences with no table entries, 4) our content plans can include the match date, if it's mentioned in the text (e.g. ``on Saturday''), 5) when we resolve a pronoun, we emit the corresponding player or team name to the content plan. With respect to Table~\ref{table:boxscore}, if the realization at the bottom was a reference summary, then by applying this process we would obtain the content plan shown in the middle of the table. On average, the plans inferred in this fashion have 59.24 table entries and 12.72 sentences.

\subsection{Content planning technical details}
\label{planning-details}

\paragraph{HiBERT.}

Conceptually, the input to HiBERT is a sequence of strings. We use three special strings, i.e., \verb+<BEG>+, \verb+<EOS>+, \verb+<EOT>+, to explicitly mark the beginning of the content plan, the end of a sentence, and the end of the plan (text), respectively. The other strings are the values from the table, e.g., \verb+Chicago_Bulls Points+, in the same order in which they appear in the text. In practice, in an attempt to leverage ROBERTA pre-training, we replace value strings with natural language sentences that we generate from each value using the templates listed in Table \ref{table:rotowire-templates}. For numeric values, such as the number of points of a team or player, similarly to \newcite{rotowire-edinlg} we compute the rank of the value among the instances of the same table entry type, and include that in the templated sentence in the form of a ``which is [1st, 2nd, 3rd, ..., Nth] best'' suffix\footnote{We use the words "which is Nth best" even when a high number is logically detrimental to the team (e.g. when it represents losses).}. With respect to the example in Table~\ref{table:boxscore}, the value \verb+Chicago_Bulls Points+ would then be represented as the natural language sentence: ``team points scored of Chicago\_Bulls is 100 which is 1st best''. 

As we did not observe a significant benefit in terms of AUC when predicting the next content plan entry on validation data, we eventually initialized our model with random weights but retained the natural language representation of the value strings. 

Because HiBERT has a sentence limit of 512, we do a pre-filtering step by discarding the table entries that are less likely to be mentioned in the summary, i.e., all player entries valued ``N/A'' and as many entries valued ``0'' as needed. 
Since the table entries aren't naturally ordered we don't feed a positional embedding $\mathbf{p}^{\mbox{sent}}_i$ in the document encoder, but we still feed it for the summary encoder.

Given the table entries and partial summary, HiBERT computes a distribution over the input sentences, where \verb+<EOS>+ corresponds to emitting a sentence break, \verb+<EOT>+ corresponds to ending the content plan, and \verb+<BEG>+ is not used.

We sample content plans from a trained model by greedy decoding with one modification: entries are not allowed to repeat in the content plan, except for sentence breaks, team names and team cities. If the highest probability sentence would have been a repeat, we instead emit the second highest, etc.

\paragraph{ETCSum.} ETC models used the same filtered set of table entries used in HiBERT as input. We concatenated these entries into a flat input sequence. Similarly, we used special strings \verb+<EOS>+, \verb+<EOT>+ and \verb+<BEG>+ which correspond to the same concepts as in HiBERT, end of sentence, end of text and beginning of text respectively. These special strings are appended at the beginning of the flat input sequence.

The partial summary input is constructed by concatenating the special string \verb+<BEG>+ and the entries that have been predicted so far, in order of prediction, with \verb+<EOS>+ indicating sentences breaks.

The full input sequence is then constructed by concatenating: a \verb+[CLS]+ delimiter, the flat input sequence, a special separator \verb+[SEP]+, the partial summary and finally a separator \verb+[SEP]+. Both the input sequence and the partial summary are padded to 6141 and 2048 respectively, adding up in total to 8192 strings for the full input, including the special delimiters.

The model uses additional inputs to construct the global-local attention. One global token is assigned to each segment in the full input, each special delimiter gets assigned a global token, as well as every sentence in the input and partial summary. The model has a maximum global token id of 512, this has to be taken into account for examples where the number of segments, input sequence sentences, special delimiters and partial inputs is larger than 512. For those examples, we don't assign global tokens to the tail of the input sequence.

To be consistent we use the same decoding strategy where we sample content plans greedily but without repeated entries allowed in the content plan except for sentence breaks, team names and team cities.

\begin{table*}[t!]
\begin{tabular}{ll}
Table entry type & Template used \\
\hline
\hline
match date & match date of match is year: YYYY month: MM day: DD day\_of\_week: W \\
\hline
team name & team name of T is V \\
team city & team city of T is V \\
TEAM-PTS\_QTR1 & team 1st quarter points of T is V \\
TEAM-PTS\_QTR2 & team 2nd quarter points of T is V \\
TEAM-PTS\_QTR3 & team 3rd quarter points of T is V \\
TEAM-PTS\_QTR4 & team 4th quarter points of T is V \\
TEAM-FT\_PCT & team free throw percentage of T is V \\
TEAM-PTS & team points scored of T is V \\
TEAM-AST & team assists of T is V \\
TEAM-LOSSES & team losses of T is V \\
TEAM-WINS & team wins of T is V \\
TEAM-REB & team rebounds of T is V \\
TEAM-TOV & team turnovers of T is V \\
TEAM-FG3\_PCT & team 3-point field goal percentage of T is V \\
TEAM-FG\_PCT & team field goal percentage of T is V \\
team playing at home or away? & T is home/away team of match \\
\hline
player first name & player first name of P is V \\
player second name & player second name of P is V \\
PLAYER-PTS & player points scored of P is V \\
PLAYER-FGM & player field goals made of P is V \\
PLAYER-FGA & player field goals attempted of P is V \\
PLAYER-MIN & player minutes played of P is V \\
PLAYER-FG3M & player 3-point field goals made of P is V \\
PLAYER-FG3A & player 3-point field goals attempted of P is V \\
PLAYER-STL & player steals of P is V \\
PLAYER-FTM & player free throws made of P is V \\
PLAYER-FTA & player free throws attempted of P is V \\
PLAYER-BLK & player blocks of P is V \\
PLAYER-AST & player assists of P is V \\
PLAYER-TO & player turnovers of P is V \\
PLAYER-PF & player fouls of P is V \\
PLAYER-REB & player rebounds of P is V \\
PLAYER-START\_POSITION & player starting position of P is V \\
PLAYER-OREB & player offensive rebounds of P is V \\
PLAYER-DREB & player defensive rebounds of P is V \\
PLAYER-FG\_PCT & player field goals percentage of P is V \\
PLAYER-FG3\_PCT & player 3-point field goals percentage of P is V \\
PLAYER-FT\_PCT & player free throws percentage of P is V \\
the team a player belongs to & P is player of T \\
\end{tabular}
\caption{The templates we use to create textual representations of the table entries. In the templates, YYYY, MMM, DD, W, T, P and V are placeholders. W encodes the day of week: Monday is 0, Sunday is 6. X is the name of a team or of a player. V is the value that the team (T) or player (P) has for the given table entry in the dataset. The names in the table entry column correspond to the names of properties in the Rotowire dataset where possible.}
\label{table:rotowire-templates}
\end{table*}

\begin{table*}[t!]
\centering
\footnotesize
\begin{threeparttable}
\begin{tabular}{l|c|ccc|c|c}
\hline
\multirow{2}{*}{Models} & \multicolumn{1}{c|}{RG} & \multicolumn{3}{c|}{CS} & CO & \multirow{2}{*}{BLEU} \\
& P\% & P\% & R\% & F1\% & DLD\% & \\ \hline
Stepwise HiBERT realized & 95.97 & 41.34 & 57.62 & 48.14 & 19.19 & 15.86 \\
Stepwise HiBERT planning only* & -- & 42.83 & 59.62 & 49.85 & -- & -- \\
Stepwise ETCSum realized & 98.78 & 45.18 & 60.14 & 51.60 & 25.87 & 17.93 \\
Stepwise ETCSum planning only* & -- & 45.53 & 60.14 & 51.82 & -- & -- \\
\hline
\end{tabular}
\end{threeparttable}
\vspace{-0.2cm}
\caption{Standard metrics for Rotowire on validation data.}
\label{table:rotowire-validation-results}
\vspace{-0.6cm}
\end{table*}

\subsection{Rotowire realization model}

The generated content plans are realized via a sequence-to-sequence transformer model initialized with ROBERTA \cite{roberta} following \cite{berts2s}, trained to emit the realization sentence by sentence. The input to the model is the concatenation of the following:
\begin{enumerate}
\item The text of the previous sentence, or the empty string (for the first sentence). (The model can use this to pronominalize team and player names if they were already introduced.)
\item The literal string \verb+" <BEG> "+ as a separator.
\item The templated realizations (cfr. Table~\ref{table:rotowire-templates}) of the entries in the sentence's content plan, space separated.
\item The literal string \verb+" <CONTEXT> "+ as a separator.
\item The templated representation of the match date.
\item For both teams, the templated representations of a) the team name, b) the team city, c) TEAM-PTS, d) TEAM-WINS, e) TEAM-LOSSES, f) whether the team was playing at home or away. These are space separated.
\item For each player \textbf{in the sentence's content plan}: the templated representations of a) PLAYER-START\_POSITION, and b) which team the player was on. These are space separated.

\end{enumerate}

The input after the \verb+" <CONTEXT> "+ separator is provided because we noticed that sometimes the content plan doesn't provide all the necessary information for realizing a sentence. For example, sometimes the target text may refer to a player by their starting position and team, which is information that wouldn't otherwise be provided to the realizer.

We create training data from the rotowire summaries and their inferred content plans by splitting them into sentences together with our inferred content plans. We realize content plans by autoregressively feeding the sentence produced in the previous step as input to the next step.

\subsection{Validation data performance}

We report performance of our best models on the Rotowire validation data in Table~\ref{table:rotowire-validation-results}.

\end{document}